\documentclass{article}
\usepackage{ijcai25}

\usepackage{times}
\usepackage{soul}
\usepackage{url}
\usepackage[hidelinks]{hyperref}
\usepackage[utf8]{inputenc}
\usepackage[small]{caption}
\usepackage{graphicx}
\usepackage{amsmath}
\usepackage{amsthm}
\usepackage{booktabs}
\usepackage{algorithm}
\usepackage{algorithmic}
\usepackage[switch]{lineno}

\usepackage{newfloat}
\usepackage{listings}

\usepackage{amsfonts,amssymb}
\usepackage{multirow}
\usepackage{booktabs}
\usepackage{amsmath}

\usepackage{makecell}

\title{Collaborative Multi-LoRA Experts with Achievement-based Multi-Tasks Loss for Unified Multimodal Information Extraction}

\author{
Li Yuan$^{1,2}$\and
Yi Cai$^{2,1}$\textsuperscript{*}\and
Xudong Shen$^{1}$\and 
Qing Li$^{3}$\and 
Qingbao Huang$^{4}$\and 
Zikun Deng$^{1,2}$\and 
Tao Wang$^{5}$
\affiliations
$^1$School of Software Engineering, South China University of Technology, Guangzhou, China \\
$^2$Key Laboratory of Big Data and Intelligent Robot (SCUT), MOE of China \\
$^3$Department of Computing, The Hong Kong Polytechnic University, Hong Kong, China\\
$^4$School of Electrical Engineering, Guangxi University, Nanning, China\\
$^5$Department of Biostatistics \& Health Informatics, King's College London, London, United Kingdom
\emails
\{seyuanli, se3048xdshen\}@mail.scut.edu.cn, \{ycai, zkdeng\}@scut.edu.cn,
csqli@comp.polyu.edu.hk, qbhuang@gxu.edu.cn, tao.wang@kcl.ac.uk
}

\begin{document}

\maketitle
\let\thefootnote\relax
\footnotetext{$^*$ Corresponding author.}

\begin{abstract}
Multimodal Information Extraction (MIE) has gained attention for extracting structured information from multimedia sources. Traditional methods tackle MIE tasks separately, missing opportunities to share knowledge across tasks. Recent approaches unify these tasks into a generation problem using instruction-based T5 models with visual adaptors, optimized through full-parameter fine-tuning. However, this method is computationally intensive, and multi-task fine-tuning often faces gradient conflicts, limiting performance.
To address these challenges, we propose collaborative multi-LoRA experts with achievement-based multi-task loss (C-LoRAE) for MIE tasks. C-LoRAE extends the low-rank adaptation (LoRA) method by incorporating a universal expert to learn shared multimodal knowledge from cross-MIE tasks and task-specific experts to learn specialized instructional task features. This configuration enhances the model’s generalization ability across multiple tasks while maintaining the independence of various instruction tasks and mitigating gradient conflicts.  Additionally, we propose an achievement-based multi-task loss to balance training progress across tasks, addressing the imbalance caused by varying numbers of training samples in MIE tasks. Experimental results on seven benchmark datasets across three key MIE tasks demonstrate that C-LoRAE achieves superior overall performance compared to traditional fine-tuning methods and LoRA methods while utilizing a comparable number of training parameters to LoRA.
\end{abstract}

\section{Introduction}
\begin{figure}    
  \centering
    \includegraphics[scale=.5]{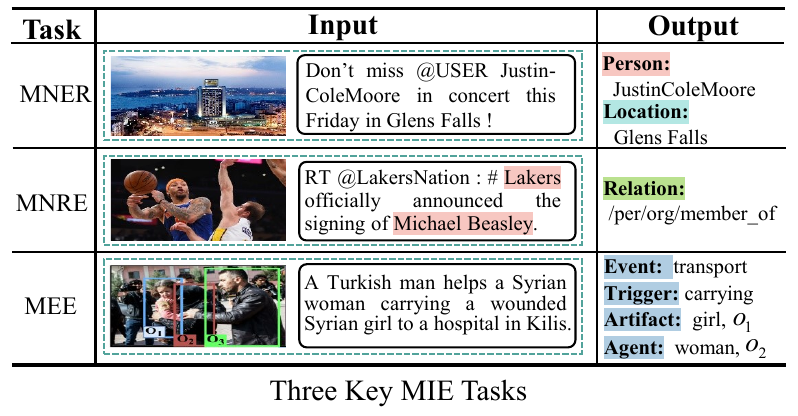}
  \caption{Three Key MIE Tasks, where O$_1$,O$_2$, and O$_3$ are visual objects.}
  \label{FIG:MIE}
  \vspace{-6mm}
\end{figure}

Recently, growing attention has been paid to multimodal information extraction (MIE) \cite{Zhang2018,li2020cross,zheng2021mnre}, which focuses on extracting structured information from multi-media sources, including text and images. Compared with text information extraction \cite{zhou2022survey}, MIE is generally a more challenging task as it requires bridging cross-modal information.  MIE commonly consists of three key sub-tasks: multimodal named entity recognition (MNER) \cite{wang-etal-2022-ita,jia2023mner}, multimodal relation extraction (MRE) \cite{zheng2021mnre,yue2023automatic}, and multimodal event extraction (MEE) \cite{li2020cross,tong2020image,10.1145}, as shown in Figure \ref{FIG:MIE}. Traditional methods only concentrate on individual tasks and employ task-specific models trained with supervised learning \cite{zheng2021mnre,li2020cross}. However,  designing, training, and maintaining individual models for each task is both time-consuming and resource-intensive. Besides,  this approach fails to efficiently leverage shared knowledge across different MIE tasks, hindering their performance and impeding the large-scale deployment of applications.

\begin{figure}   
  \centering
    \includegraphics[scale=.46]{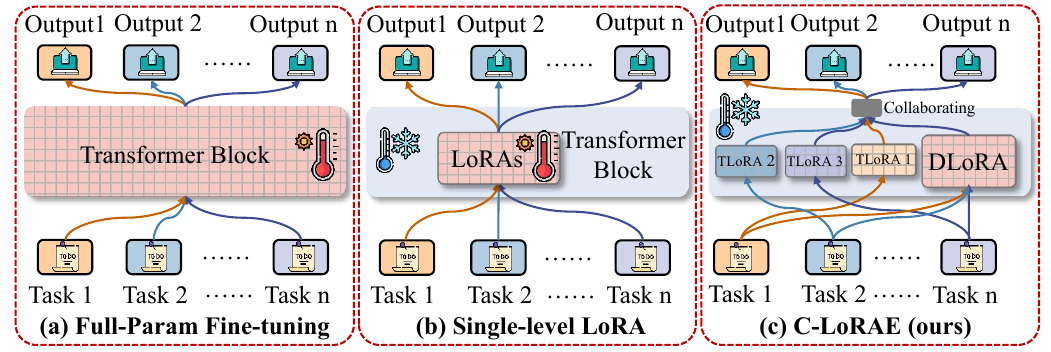}
  \caption{Different fine-tuning methods for MIE.}
  \label{FIG:1}
  \vspace{-6mm}
\end{figure}

To address the various MIE tasks in a unified way to share
Information across different MIE tasks, \cite{Sun2024} proposed an approach called the Unified Multimodal Information Extractor (UMIE). This approach converts the training objectives of different MIE tasks into generation problems and utilizes a T5 model with a visual adaptor to generate outputs for various MIE tasks by full-parameter fine-tuning (FPFT) with multi-task instruction learning. However, this method has two key issues that limit its practical application and performance. (1) FPFT necessitates significant training time and hardware resources (e.g., GPUs) \cite{li2021prefix}, as illustrated in Figure \ref{FIG:1}(a).
This approach challenges adapting to larger language models with limited hardware. (2) Multi-task instruction learning often aggregates diverse task data indiscriminately, leading to suboptimal performance \cite{dai2024cotbal,chen2024llava}, known as \emph{negative transfer} \shortcite{zhang2021survey,Yun2023}. Variations in instruction formats and knowledge domains across tasks can cause gradient conflicts during simultaneous training.

An effective approach to address the first problem caused by full-parameter fine-tuning involves utilizing parameter-efficient fine-tuning methods (PEFT) \cite{li2021prefix}, such as LoRA \shortcite{hu2021lora}. By freezing the pre-trained model weights, LoRA reduces the number of trainable parameters in the Transformer.
Specifically, it involves training only a pair of low-rank decomposition weight matrices for each linear layer. This strategy has been widely adopted in fine-tuning large language models (LLMs) and large vision-language models (LVLMs) \cite{liu2023visual,yin2024lamm}. 
Recent studies \cite{wu-etal-2024-mixture-subspaces,dou2024loramoe} have integrated the Mixture of Experts (MoE) \shortcite{yuksel2012twenty} concept into LoRA, using multiple LoRA modules as distinct experts to improve the model's capacity to apply world knowledge for downstream tasks, as shown in Figure \ref{FIG:1}(b). However, the limitation of \emph{negative transfer} persists, as single-level LoRA-based approaches typically mix all instruction tasks without addressing potential instruction conflicts, thereby limiting the model's performance.

\begin{figure*}    
  \centering
    \includegraphics[scale=.57
    ]{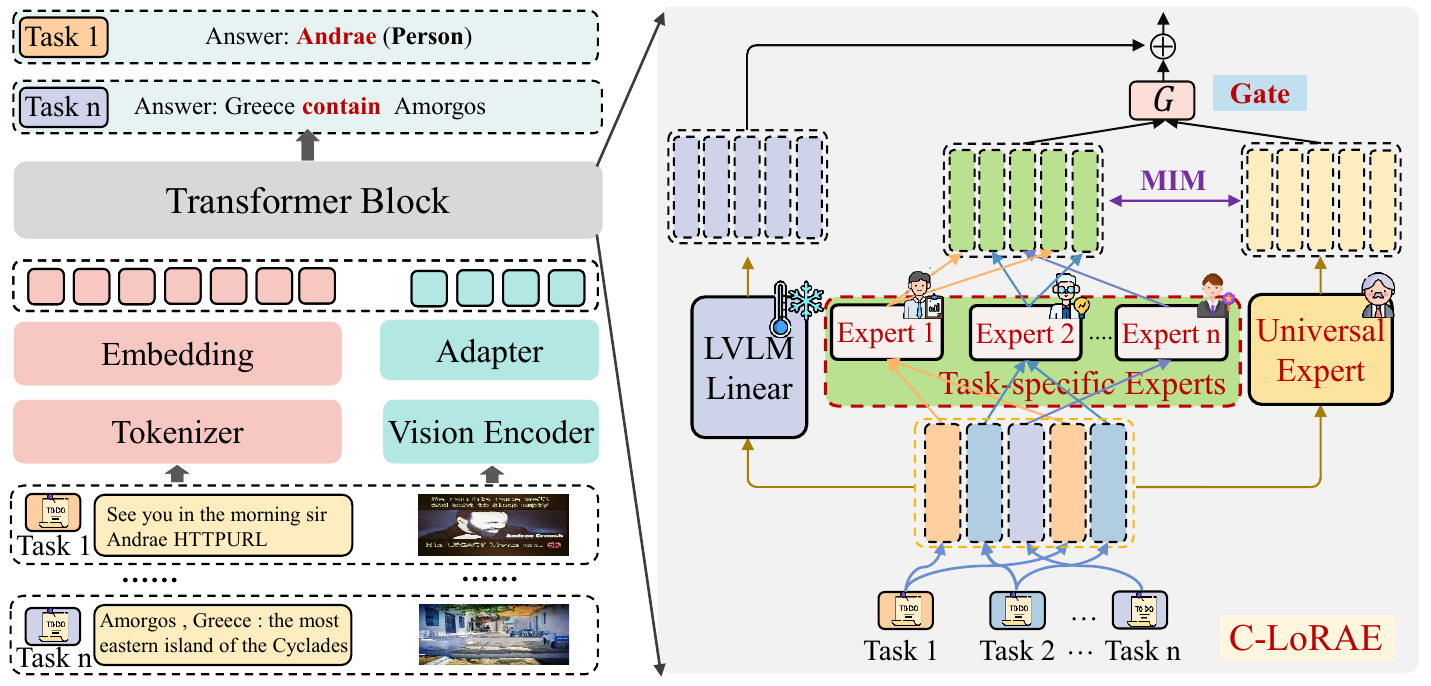}
  \caption{Overall framework of our Collaborative Multi-LoRA Experts. Our model is based on the General Large Visual-Language Model. The input text is tokenized and embedded and then concatenated with the visual input to feed into the LLMs.}
  \label{FIG:3}
\end{figure*}

To facilitate knowledge sharing among MIE tasks while minimizing conflicts arising from integrating diverse instructional tasks within a lower training budget, this study proposes Collaborative Multi-LoRA Experts (C-LORAE) with an achievement-based multi-task loss for MIE tasks. As illustrated in Figure \ref{FIG:3}, we extend the vanilla LoRA in three ways. First, inspired by the MOE framework \cite{jacobs1991adaptive}, we introduce a universal LoRA expert along with a set of task-specific experts. This approach assimilates shared multi-model information across various instruction tasks while preserving task-specific knowledge. This configuration enhances the model's generalization ability across multiple tasks while reducing multi-task gradient conflicts. Second, we utilize mutual information maximization to facilitate the effective exchange of information between task-specific experts and the universal expert. Additionally, to enable autonomous and efficient collaboration between the task-specific experts and the universal expert, we propose an expert-motivated gate router that determines whether the features of each token should be output by the universal expert or a task-specific expert. This mechanism allows the model to leverage both generalization and specialization simultaneously.

Moreover, since the number of training samples for individual MIE tasks varies significantly, a multi-task model can easily become biased toward the task with the larger number of training samples \cite{Yun2023}, leading to suboptimal overall performance. Therefore, we use achievement-based multi-task loss, which is derived from the current performance across different tasks, to dynamically balance the training progress across instruction tasks and task-specific experts.
Our main contributions can be summarized as follows:

\begin{itemize}
\item We propose a C-LoRAE method to address MIE tasks, consisting of a universal LoRA expert and a set of task-specific LoRA experts. This approach facilitates knowledge sharing across MIE tasks while resolving gradient conflicts that arise when instruction fine-tuning a large language-vision model on diverse MIE datasets.

\item 
We propose an achievement-based multi-task loss to address the training imbalance caused by varying numbers of training samples in MIE tasks. This loss function balances the training progress of our proposed C-LoRAE model by considering the current performance across different MIE instruction tasks.

\item  We conducted experiments across seven datasets from MIE tasks. The results demonstrate that our proposed model, C-LoRAE, which utilizes significantly fewer training parameters compared to full fine-tuning, achieves superior overall performance across most tasks and establishes state-of-the-art performance on four datasets. Moreover, C-LoRAE consistently outperforms fine-tuning with vanilla LoRA across all datasets.

\end{itemize}

\section{Method}
In this section, we first introduce the collaborative multi-LoRA experts to replace the vanilla linear layers in  LVLMs, as depicted in Figure \ref{FIG:3}. Then, we propose an achievement-based multi-task loss to address the training imbalance caused by varying numbers of training samples in MIE tasks.
\subsection{Problem Formulation}

As illustrated in Figure \ref{FIG:3}, a Transformer-based LVLM, such as UMIER \cite{Sun2024} and LLaVA \cite{liu2023visual}, can be structured to address MIE in a unified way as follows:
\begin{eqnarray}\label{eq:1}
T_k^{ans} = {f_{{\rm{LVLM}}}}({f_{vis}}({I_k}),{f_{emb}}(T_k^{in}))
\end{eqnarray} 
where $f_{emb}$ denotes the transformation of input text $T^{in}$ into a word embedding matrix, and $f_{vis}$ represents the visual encoder with a visual adapter to convert an image into a sequence of visual embeddings. The triplets $(T_k^{in},{I_k},T_k^{ans})_{k = 1}^K$ represent the training data for the $k$-th training sample among all $K$ samples.

\subsection{Preliminary} \textbf{Low-rank Adaptation (LoRA)} is an efficient method for fine-tuning LLMs and LVLMs. Given that our proposed approach builds on the foundational principles of LoRA, it is crucial to provide an overview of this method.
Formally, a linear layer $h = {W_0}x$ with input $x \in \mathbb{R}{^{{d_{in}}}}$ and  weight matrix ${W_0} \in \mathbb{R}{^{{d_{out}} \times {d_{in}}}}$. In this setup, the LoRA module comprises an original linear layer from LVLMs with a LoRA layer decomposition, which is defined as follows:
\begin{eqnarray}\label{eq:1_2}
h = LoRA(x) = {W_0}x + BAx
\end{eqnarray} 
where $ A \in \mathbb{R}{^{r \times {d_{in}}}}$ and $B \in \mathbb{R}{^{{d_{out}} \times r}}$ are the low rank matrices. Since \( r \ll \min(d_{out}, d_{in}) \), the number of trainable parameters in \( A \) and \( B \) is significantly smaller than that in \( W_0 \). During the training of a LoRA module, only the matrices \( A \) and \( B \) are updated.
\subsection{Collaborative Multi-LoRA Experts}
Our method aims to facilitate knowledge sharing among MIE tasks while reducing gradient conflicts arising from mixing diverse instruction tasks. To achieve this, we propose collaborative multi-LoRA experts: a universal expert dedicated to learning shared knowledge information from all tasks, alongside a set of task-specific LoRA experts designed for learning task-specific features within individual tasks. Moreover, we employ mutual information maximization to facilitate the exchange of information between task-specific experts and the universal expert. Finally, we propose an experts-motivated gate router to obtain final token representations, which are customized for each token from different tasks. This router also helps to determine whether to utilize the universal LoRA expert or the task-specific LoRA for each token.

\subsubsection{Universal LoRA Expert} To maintain knowledge exchange among MIE tasks and enhance the model's generalization ability by training on extensive instructions \cite{wei2021finetuned}, we propose a universal LoRA expert (ULoRA). ULoRA adopts the vanilla LoRA framework and is capable of learning universal multimodal representations across various tasks. Formally, given the \emph{i}-th training sample in \emph{n} instruction task ${x^n_{i}}$, the output of the universal expert $U_i^n$ is defined as:
\begin{eqnarray}\label{eq:2}
U_i^n = ULoRA({x^n_{i}})
\end{eqnarray} 
where the trainable matrices from the universal expert are denoted as  $A^U \in \mathbb{R}^{{r \times {d_{in}}}}$ and $B^U\in \mathbb{R}^{{ {d_{out}\times r}}}$. Each element ${x^n_{i,j}}$ in ${x^n_{i}}$ represents the \emph{j}-th token of the sample.

\subsubsection{Task-specific LoRA Experts}

To address the issue of \emph{negative transfer} caused by gradient conflicts across MIE tasks, we introduce a set of task-specific experts that are optimized solely by their respective task datasets. Specifically, for each instruction task, we designate a task-specific expert, resulting in \emph{N}=3 (MNER, MRE, and MEE) task-specific experts in the MIE task framework. Assuming that the \emph{i}-th training sample belongs to the instruction task \emph{n}, the task-specific LoRA expert $T{^n}LoRA$ outputs $D{^n_{i}}$ of ${x^n_{i}}$ are as follows:
\begin{eqnarray}\label{eq:4}
D{^n_{i}} = T{^n}LoRA({x^n_{i}})
\end{eqnarray} 
Considering that each task-specific LoRA expert fine-tunes solely on its dataset and thus has less data for learning compared to the universal LoRA expert, we define the trainable matrices for the \emph{n}-th task-specific LoRA expert as $A_n^D\in\mathbb{R}^{\frac{r}{N} \times {d_{in}}}$ and $B_n^D \in \mathbb{R}^{{{d_{out}\times \frac{r}{N}}}}$. This approach aims to minimize the number of training parameters while ensuring that the model retains the capability to learn task-specific features.

\subsubsection{Collaborating of Experts}
\textbf{(1) Mutual Information Maximization} To harmonize task-specific experts and the universal expert, we regard the universal expert as a teacher who imparts essential universal knowledge to students in various fields, enhancing their ability to model the bridge between text and image \cite{Ahn2019}. We employed the variational information maximization method to maximize the mutual information (MIM) between universal expert representations $U_i^n$ and task-specific expert representations $D_i^n$, which is defined as:
\begin{eqnarray}\label{eq:5}
{{\cal L}_{{\rm{MIM}}}} = \sum\limits_{n \in N} {\sum\limits_{i \in |n|} {{\rm{MIM}}(D_i^n,U_i^n)} }
\end{eqnarray} 
where $|n|$ denotes the number of samples in task $n$, and ${\cal L}_{{\rm{MIM}}}$ the loss function for optimizing mutual information. MIM(·) refers to the specific MIM method used in the optimization.
\newline
\textbf{(2) Experts-motivated Gate Router} 
To tailor the contributions of the task-specific expert representation $D_i^n$ and the universal expert representation $U_i^n$ across various tasks, we introduce an expert-motivated gate router. For each input token $x_{i,j}^n$, the router $\mathcal{G}$ learns to activate the most suitable weights from the universal expert and the \emph{n}-th task expert. The calculation of the gate router is defined as follows:
\begin{eqnarray}\label{eq:6}
{{\cal G}^n_{i,j}}{\rm{ = softmax}}({W_g}({{x}^n_{i,j}}))
\end{eqnarray} 
where ${x}^n_{i,j}$ represent the \emph{j}-th token of the input sequence $x^n_{i}$ and $G_{i,j}^n = \{ g_{i,j}^{n,1};{g}_{i,j}^{n,2}\}$ denotes the gate weights for the universal expert and \emph{n}-th task-specific expert. Additionally, ${W_g}\in\mathbb{R}^{2 \times d_{out}}$ is shared among all the experts, and represents the linear gate weights. Thus, the output of the collaborative multi-LoRA experts $h^n_{i,j}$ is defined as follows:
\begin{eqnarray}\label{eq:7}
h_{i,j}^n =  g_{i,j}^{n,1} * U_{i,j}^n+g_{i,j}^{n,2} * D_{i,j}^n
\end{eqnarray} 
We use a weighted sum of the collaborative multi-LoRA experts, $h_{i,j}^n$, and the original linear layer on input $x_{i,j}^n$ to compute final output $\widetilde y_{i,j}^n$, which can be expressed as follows:
\begin{eqnarray}\label{eq:8}
\widetilde y_{i,j}^n = {W_0}(x_{i,j}^n) + \frac{\alpha }{r}h_{i,j}^n
\end{eqnarray} 
where $\alpha$ is a hyper-parameter to facilitate tuning trainable low-rank matrices \emph{r}.
\begin{table}[]
\centering
\small
\begin{tabular}{c|cccc}

\toprule
\textbf{Task} & \textbf{Dataset}  & \textbf{Train}  & \textbf{Dev}   & \textbf{Test}  \\
\hline
\multirow{3}{*}{MNER} & Twitter-15 & 4,000  & 1,000 & 3,257 \\

                      & Twitter-17 & 2,848  & 723   & 723   \\
                      & SNAP       & 3,971  & 1,432 & 1,459 \\
\hline
\multirow{2}{*}{MRE}  & MRE-V1    & 7,824  & 975   & 1,282 \\
                      & MRE-V2    & 12,247 & 923   & 832   \\
\hline
\multirow{3}{*}{MEE}  & ACE2005    & 4424   & 224   & -     \\
                      & SWiG       & 16451  & 3350  & -     \\
                      & S2E2       & -      & -     & 309  \\
 \bottomrule
\end{tabular}
\caption{The statistics of six MIE datasets.}
\vspace{-4mm}
\label{statistic}	
\end{table}

\begin{table*}[ht]
\centering
\small
\setlength{\tabcolsep}{0.3mm}
\begin{tabular}{c|ccc|cc|cc|c}
\toprule
\multirow{2}{*}{\textbf{Model}} & \multicolumn{3}{c|}{MNER}       & \multicolumn{2}{c|}{MRE} & \multicolumn{2}{c}{MEE} & \multicolumn{1}{|c}{\multirow{2}{*}{All}} \\
& Twitter-15 & Twitter-17 & SNAP & MRE-V1    & MRE-V2    & ${\rm{MED}}$    & ${\rm{MEAE}}$  & \\
\hline
UMT(\citeauthor{yu2020})                      & 73.4       & 73.4       & -    & -          & 65.2       & -          & -        &-  \\
UMGF(\citeauthor{zhang2021multi})                    & 74.9       & 85.5       & -    & -          & -          & -          & -       &-  \\
MEGA(\citeauthor{Zheng2021})                   & 72.4       & 84.4       & 66.4 & -          & -          & -          & -      &-    \\
RpBERT(\citeauthor{sun2021rpbert})                  & 74.4       & -          & 85.7 & -          & -          & -          & -        &- \\
R-GCN(\citeauthor{zhaomm2022})                   & 75.0       & 87.1       & -    & -          & -          & -          & -        &-  \\

ITA(\citeauthor{wang-etal-2022-ita})                     & \textbf{78.9}       & 89.8       & 90.2 & -          & 66.9       & -          & -       &-   \\
MoRe(\citeauthor{wangemnlp2022})                    & 77.3       & 88.7       & 89.3 & -          & 65.8       & -          & -       &-  \\      
HVPNeT(\citeauthor{chen2022good})                  & 75.3       & 86.8       & -    & 81.8       & \underline{81.9}          & -         & -        &-  \\
MKGFormer(\citeauthor{chen2022hybrid}) & -           &87.5  & -          &-      &\textbf{82.0}      &-   &-   \\    

MNER-QG(\citeauthor{jia2023mner})                 & 75.0       & 87.3       & -    & -          & -          & -          & -        &- \\
HamLearning(\citeauthor{liu2023novel})             & 76.5       & 87.1       & -    & -          & -          & -          & -       &-  \\
EviFusion(\citeauthor{liu2023integrating})               & 75.5       & 87.4       & -    & -          & -          & -          & -        &- \\
BGA-MNER(\citeauthor{chen2023chain})                & 76.3       & 87.7       & -    & -          & -          & -          & -        &- \\
WASE(\citeauthor{li2020cross})                    & -          & -          & -    & -          & -          & 50.8       & 19.2     &- \\
Unicl(\citeauthor{10.1145})                   & -          & -          & -    & -          & -          & 57.6       & 23.4     &-\\

\hline
UMIER$_{Base}$(\citeauthor{Sun2024}) $\dag$       & 76.1     & 88.1       & 87.7 & 84.3       & 74.8       & 60.5       & 22.5      &494.0 \\

UMIER$_{Large}$        & 77.2       & \underline{90.7}       & \underline{90.5} & \underline{85.0}       & 75.5       & \underline{61.0}       & \underline{23.6}      &503.5 \\
UMIER$_{XLarge}$      & \underline{78.2}       & \textbf{91.4}       & \textbf{91.0} & \textbf{86.4}       & 76.2       & \textbf{62.1 }      & \textbf{24.5}     &509.8 \\
\hline

C-LoRAE$_{Base}$    & 74.4$ \downarrow\!\!(1.9)$       & 86.3$\downarrow\!\!(1.8)$         & 85.8$\downarrow\!\!(1.9)$  & 87.3$\uparrow\!\!(3.0)$      & 81.4$\uparrow\!\!(6.6)$     & 60.7$\uparrow\!\!(0.2)$       & 28.4$\uparrow\!\!(5.9)$    &503.9$\uparrow\!\!(9.9)$   \\
C-LoRAE$_{Large}$    & 75.9$\downarrow\!\!(1.3)$       & 88.9$\downarrow\!\!(1.8)$         & 89.3$\downarrow\!\!(1.2)$  & \underline{89.2}$\uparrow\!\!(4.3)$     & \underline{82.8}$\uparrow\!\!(7.3)$      & \textbf{64.5}$\uparrow\!\!(3.5)$      & \underline{31.2}$\uparrow\!\!(7.6)$     &521.3$\uparrow\!\!(17.8)$   \\
C-LoRAE$_{XLarge}$   & 77.8$\downarrow\!\!(0.4)$       & 89.8$\downarrow\!\!(1.6)$      & \underline{90.5}$\downarrow\!\!(0.5)$ & \textbf{89.6}$\uparrow\!\!(3.2)$       & \textbf{83.2}$\uparrow\!\!(6.0)$       & \underline{63.5}$\uparrow\!\!(1.4)$       & \textbf{32.6}$\uparrow\!\!(8.1)$  &527.0$\uparrow\!\!(17.2)$  \\     
\bottomrule
\end{tabular}
\caption{The comparative test results for three multimedia information extraction tasks are based on the F$_1$ score. All values represent the total score across all tasks. The best performance is highlighted in bold, while the second-best is underlined. For the baseline, we used the results from the \protect\cite{Sun2024} report. Models designed for specific tasks may lack generalizability; results that were not reported are marked with  ``-".
}
\label{table:main}
\end{table*}

\subsection{Achievement-based Multi-task Loss}
Since the numbers of training samples for individual MIE tasks can significantly differ, the multi-task models can be easily biased toward the dominant task, which has larger numbers of training samples \cite{Yun2023}. Thus, to balance training progress across instruction tasks in C-LoRAE, we propose an achievement-based multi-task loss. This approach, originally inspired by focal loss, was introduced to address the class imbalance in object detection \cite{lin2017focal}. Formally, we define the achievement $p_c$ of each task as the ratio of the current and single-task metric $F_1$ score of the state-of-the-art model. Thus, the achievement-based weight can be defined as follows:
\begin{eqnarray}\label{eq:9}
w_t^m = {(1 - \frac{{s_t^m}}{{\partial  \cdot {p_m}}})^\gamma }
\end{eqnarray} 
where $w_t^m$ and $s_t^m$ represent the weight and $F_1$ score of the $m$-th dataset at the $t$-th training epoch, respectively. The $P_m$ denotes the SoTA results of the $m$-th dataset. However, task weights can unintentionally increase if the $F_1$ score of the multi-task model surpasses that of single-task models. To prevent this unintended increase and encourage the model to perform above the SoTA, we introduce a slight margin $\partial > 1$. Reducing task weights is analogous to theoretically decreasing the learning rate, which could lead to underfitting in the corresponding tasks. To mitigate the risk of underfitting, we normalize task weights using $softmax$.

To measure the ground truth distribution and the predicted tagging distribution of the $m$-th task at the $t$-th training epoch, we employ the cross-entropy error $\mathcal{L}_m^t$. Then, we use $\mathcal{L}_m^t$ with their corresponding weight $w_t^m$ to compute the multi-task loss $\mathcal{L}_{MT}$. Finally, we jointly optimize the multi-task loss and the loss of MIM. Therefore, the final jointly optimized objective is computed as follows:
\begin{eqnarray}
\nonumber
&& {\cal L} = {{\cal L}_{MT}} + \beta  \cdot {{\cal L}_{{\rm{MIM}}}}\\
&& {{\cal L}_{MT}} = \sum\limits_{t = 1}^T {\sum\limits_{m = 1}^M {w_t^m \cdot {\cal L}_m^t} } \\
&& {\cal L}_m^t = \sum\limits_{({x^m},{y^m}) \in {D^m}} {CE(p({{\widetilde y}^m}|{x^m},{y^m};\theta),{y^m})} \nonumber
\end{eqnarray}
where $\beta$ is a hyperparameter that regulates the influence on optimizing the MIM term, while $\theta$ represents the trainable parameters. $D^m$ denotes the complete sample set of the $m$-th task dataset.

\section{Experiments}
We conducted extensive experiments on various MIE datasets to evaluate the effectiveness and efficiency of the proposed methods for MIE. Additional details and supplementary experimental results are provided in the Appendix.

\subsection{Implementation Details}
All models utilized the Adam optimizer with a learning rate of 1e-4 and a decay factor of 0.5. During training, experiments were conducted on four NVIDIA RTX 3090 GPUs or GPUs of equivalent performance. The maximum text input length was set to 256, and training lasted for 20 epochs. To accommodate GPU memory constraints, different batch sizes were used: 10 for FLAN-T5-base, 4 for FLAN-T5-large, and 2 for T5-XLarge on each GPU. We configured the number of task-specific LoRA experts to 3 to match the MIE instruction task number and the influence on optimizing the MIM term $\beta$ is 0.01. Detailed discussions on the low-rank dimensionality and dropout rate in our proposed C-LoRAE will be provided in subsequent sections.

 \begin{figure}   
  \centering
    \includegraphics[scale=0.22]{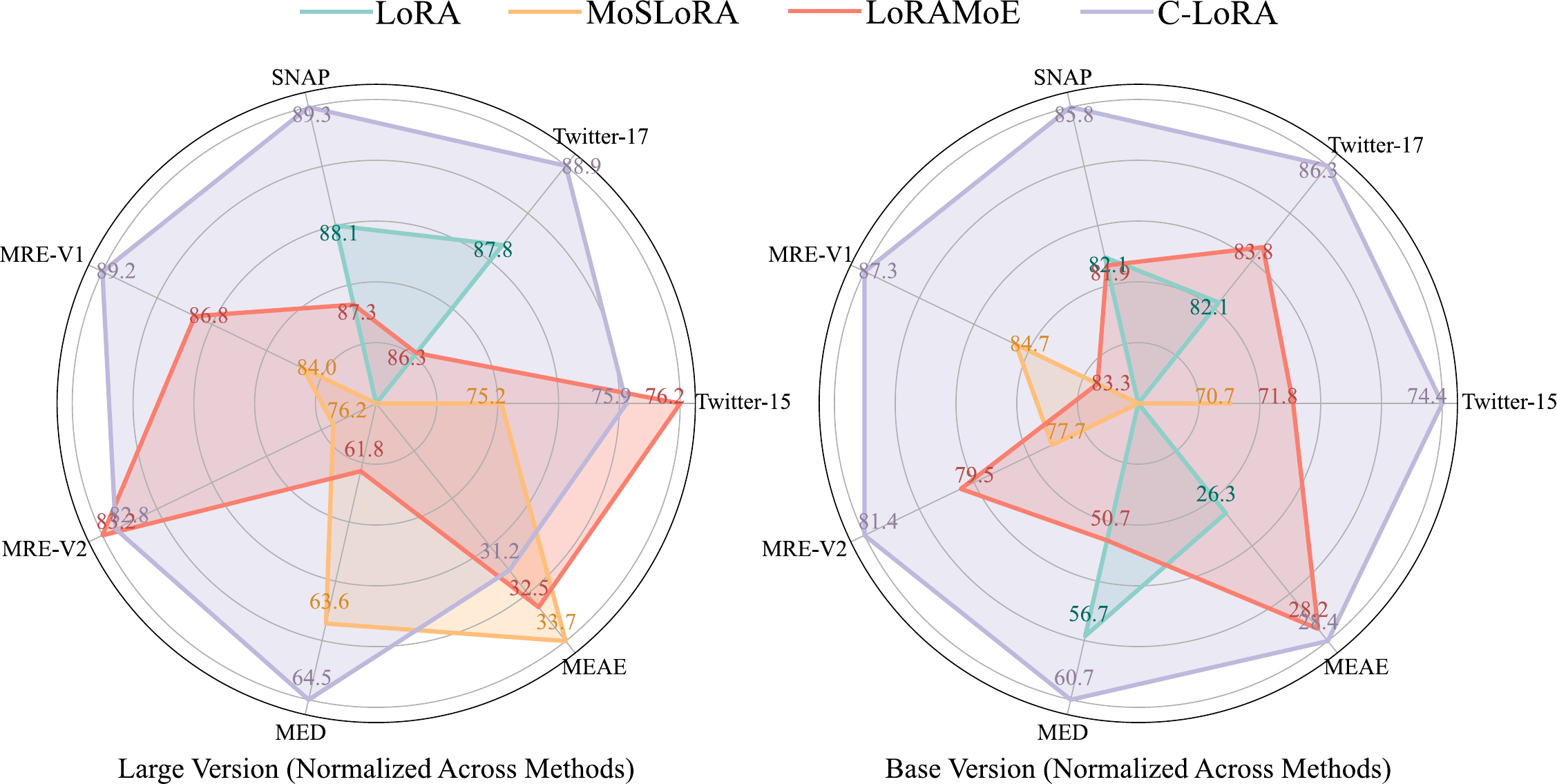}
  \caption{Normalized performance across seven tasks for LoRA, MoSLoRA, LoRAMoE, and our proposed C-LoRA.}
  \label{FIG:radar}
\end{figure}
\subsection{Dataset Details}
Following the experimental setup in \cite{Sun2024}, we train and evaluate the proposed C-LoRAE method on MNER, MRE, and MEE tasks, 
\begin{itemize}
\item \textbf{MNER} 
Following previous studies, we used the Twitter-15 dataset \cite{Zhang2018}, the SNAP dataset \cite{lu-etal-2018-visual}, and the Twitter-17 dataset \cite{yu2020}. These datasets are sourced from the social media platform and are commonly employed in prior research.
\item \textbf{MRE} 
We adopt the MRE data set \cite{zheng2021mnre} constructed from social media, which is currently the only available MRE data set. This data set is divided into versions V1 and V2, with V2 being the refined version of V1.
\item {\textbf{MEE} Follow the previous task setting \cite{tong2020image}, we train our proposed method using datasets such as ACE2005 and SWiG, and subsequently evaluate its performance using the ${{\rm{M}}^{\rm{2}}}{{\rm{E}}^{\rm{2}}}$ dataset.}
\end{itemize}
To ensure uniform representation across all tasks, we reformat all datasets into a standardized JSON format. Detailed statistics for each dataset are presented in Table \ref{statistic}.

\begin{table*}[]
\centering
\small
\setlength{\tabcolsep}{0.8mm}
\begin{tabular}{c|ccc|cc|cc|cc}
\toprule
\multirow{2}{*}{\textbf{Model}} & \multicolumn{3}{c|}{MNER}       & \multicolumn{2}{c|}{MRE} & \multicolumn{2}{c}{MEE} & \multicolumn{1}{|c}{\multirow{2}{*}{All}} & \multicolumn{1}{c}{\multirow{2}{*}{TP}} \\
& Twitter-15 & Twitter-17 & SNAP & MRE-V1    & MRE-V2    & ${\rm{MED}}$    & ${\rm{MEAE}}$  & \\
\hline

C-LoRAE$_{Large}$  & 75.9      & 88.9       & 89.3 & 89.2       & 82.8       & 64.0       & 31.2   &521.3 &39M \\

\hline
UMIER$_{Large}$(\emph{Only} ULoRA)    & 74.5      & 87.8       & 88.1 & 82.1       & 74.9      & 61.0       & 25.3 &493.7 &19M  
\\
UMIER$_{Large}$(\emph{Only} TLoRA)    & 72.9      & 85.2       & 82.3 & 85.8       & 81.2       & 62.4       & 29.5  &499.3  &19M
\\
C-LoRAE$_{Large}$(\emph{w/o} EGR)    & 74.9     & 87.8      & 88.4 & 86.4       & 80.8       & 61.8      &  27.8  &507.9 &38M    \\

C-LoRAE$_{Large}$(\emph{w/o} AML)    & 74.3    & 87.2       & 86.8 & 87.6       & 81.7       & 63.3       & 29.8     &510.7 &39M  \\

C-LoRAE$_{Large}$(\emph{w/o} MIM)    & 75.4     & 88.2      & 88.1 & 87.1     & 82.0       & 63.8      & 30.2     &514.8  &39M \\
\bottomrule
\end{tabular}
\caption{
An ablation study was conducted on the test F$_1$ scores using the large version of the model. TP denotes the number of trainable parameters during optimization, $Only$ indicates the exclusive use of a specific module, and $w/o$ signifies the exclusion of a module.}
\label{table: ablation}
\end{table*}


\subsection{Main Results}
As depicted in Table \ref{table:main}, we observe that UMIER shows improvement over most previous independent task models, particularly as the scale of Flan-T5 increases, resulting in consistent performance enhancements. Although UMIER exhibits capabilities close to the SoTA method (ITA) in the MNER task, there is a significant performance gap in the MRE V2 task compared to the MRE SoTA (MKFGformer). Even with the largest model, UMIER$_{XLarge}$, there is a decrease of 5.6\% in terms of F$_1$ score in the MRE-V2 dataset. One possible explanation is the 
\emph{negative transfer} caused by fine-tuning the full-parameter model on various instruction tasks, leading the model to fail to achieve optimal performance on MRE tasks.

We replaced the vanilla linear layer of the Transformer in the UMIER model with the proposed C-LoRAE structure, denoted as C-LoRAE. 
Despite having a limited number of trainable parameters, our model achieves comparable, and in many cases superior, performance relative to the original UMIER model. Specifically, for the "All" metric, which aggregates $F_1$ scores across all datasets to measure overall performance, C-LoRAE consistently and significantly outperforms the original UMIER model. Notably, in the \emph{Large} and \emph{XLarge} versions, C-LoRAE demonstrates improvements of 17.8 and 17.2 points, respectively. These results suggest that the proposed method facilitates knowledge sharing across MIE tasks while reducing conflicts arising from the integration of diverse instruction data types. Besides, Our model shows substantial improvements in the MRE and MEE tasks compared to UMIER. The \emph{XLarge} version sets a new SOTA across four datasets, indicating that the MRE and MEE tasks benefit significantly from our collaborative multi-LoRA experts approach, especially in comparison to the MNER task.

We further compared the performance of the proposed method with current mainstream efficient fine-tuning approaches, as shown in Figure \ref{FIG:radar}. The results demonstrate that our method outperforms the single-stage LoRA approach across most tasks, with a particularly pronounced advantage on the T5-Base backbone. This can be attributed to the dual-stage LoRA structure in our framework, which effectively learns general knowledge while significantly reducing conflicts between tasks.

\subsection{Ablation Study}
To investigate the effectiveness of different components in the C-LoRAE structure, we conduct ablation studies on the universal LoRA expert (ULoRA), task-specific LoRA experts (TLoRA), mutual information maximization (MIM) module, and the experts-motivated gate router (EGR). Additionally, we evaluate the effect of the proposed achievement-based multi-task loss (AML).

The results are reported in Table \ref{table: ablation}. First, we observe that employing \emph{only} ULoRA, which utilizes the standard LoRA for fine-tuning UMIER, results in a performance decrease of 27.6 in the ALL metric. Notably, the MRE-V1, MRE-V2, and MEE datasets exhibit significant declines compared to others. This indicates that using a single LoRA for fine-tuning all instructional tasks does not consistently improve performance, particularly when gradient conflicts among these tasks are present. Subsequently, we implemented task-specific LoRA for each instruction task (\emph{only} TLoRA). This approach led to a notable performance decline in the MNER task compared to \emph{only} ULoRA, primarily due to the MNER task having less available training data than the others. Additionally, ULoRA struggles to effectively leverage shared knowledge from other tasks, which limits its performance enhancement capabilities.

Thirdly, excluding EGR (\emph{w/o} EGR) and replacing it with element-wise addition results in significant performance degradation across most sub-tasks. This suggests that the importance of expert collaboration varies among different tasks, and simple addition cannot effectively manage diverse instructional tasks. Finally, the AML strategy successfully balances performance across various instructional tasks, resulting in overall performance improvements. Moreover, while the model remains functional without the MIM module, its absence reduces overall performance.

\begin{table}[]
\small
\begin{tabular}{c|cc}
\toprule
\textbf{Method}         & \textbf{Twitter-17} & \textbf{MRE-V2} \\
\hline
UMT(\citeauthor{yu2020})           & 60.9       & -       \\
UMGF(\citeauthor{zhang2021multi})              & 60.9       & -       \\
FMIT(\citeauthor{lu2022flat})           & 64.4       & -       \\
CAT-MNER(\citeauthor{9859972})      & 64.5       & -       \\
BGA-MNER\citeauthor{chenandliu2023}       & 64.9       & -       \\
ChatGPT $\ddag$        & 57.5       & 35.2    \\
GPT-4   $\ddag$       & 66.6       & 42.1    \\
\hline
UMIER$_{Base}$       & 66.8       & 67.3    \\
UMIER$_{Large}$     & 68.5       & 68.8    \\
\hline
C-LoRAE$_{Base}$   &    66.2         &    76.5     \\
C-LoRAE$_{Large}$  &   67.8         &  78.2        \\
\bottomrule
\end{tabular}
\caption{Comparison of the generalization ability. The ChatGPT$\ddag$ and GPT-4$\ddag$ result in in-context learning setting come from \protect\cite{chen2023chain}
.}
\label{table:gener}
\end{table}

\subsection{Generalization Analysis}
To investigate the generalization capability of the C-LoRAE structure, we followed the settings of previous work \cite{9859972,Sun2024} and evaluated its performance on the Twitter-17 and MRE-V2 datasets. Specifically, we excluded these datasets and only used the remaining training data in model training.

 Table \ref{table:gener} presents the results. We observe that the proposed C-LoRAE maintains strong generalization ability, despite utilizing fewer trainable parameters. 
Considering that generalization has also been verified in MEE, these experiments suggest that the proposed structure can effectively adapt to and generalize across new datasets, even without direct training on those specific datasets. The results demonstrate that C-LoRAE not only mitigates conflicts arising from mixing diverse instruction data types through task-specific LoRA but also benefits from the Universal LoRA, which enables knowledge sharing across all datasets. Furthermore, C-LoRAE effectively learns from multimodal data sources and successfully transfers knowledge across various tasks.

\begin{table}[]
\setlength{\tabcolsep}{0.8mm}
\centering
\small
\begin{tabular}{ccccc}
\toprule
\multicolumn{2}{c}{\textbf{Methods}}             & \textbf{TP} & \textbf{\makecell{fwd\\ FLOPs  (G)}} & \textbf{\makecell{fwd+bwd\\ FLOPs (G)}} \\
\hline

\multirow{5}{*}{\makecell{UMIE\\(Base)}} & Full Fine-tuning & 287.0       & 21.3                   & 63.8                       \\
& Vanilla LoRA     & 15.5        & 22.0                   & 65.9                       \\
 & MoSLoRA          & 15.8        & 26.3                   & 78.9                       \\
& LoRAMoE          & 30.5        & 22.7                   & 68.2                       \\
 & C-LoRA(\emph{ours})     & 30.4        & 23.2                   & 69.5        \\
\hline
 \multirow{5}{*}{\makecell{UMIE\\(Large)}}  & Full Fine-tuning & 843.0         & 73.2                   & 219.0                      \\
& Vanilla LoRA     & 20.3       & 74.1                   & 223.4                      \\
& MoSLoRA          & 20.4        & 89.1                   & 267.3                      \\
 & LoRAMoE          & 40.0        & 75.2                   & 225.5                      \\
& C-LoRA(\emph{ours})     & 39.9        & 75.7                  & 227.2                      \\
\toprule
\end{tabular}
\caption{The sensitivity analysis assesses the different models. TP represents the number of trainable parameters during optimization, while \textit{fwd} and \textit{bwd} indicate a single forward and backward process, respectively. G represents the GFLOPs.}
\label{sensitive parameter}	
\end{table}

\subsection{Sensitivity Analysis}
In this section, we analyze the sensitivity of the proposed C-LoRA under the \emph{Base} (rank = 128) and \emph{Large} (rank = 64) versions. For comparison, we evaluate the vanilla UMIER, MoSLoRA, and LoRAMoE. We examined the floating-point operations (FLOP) with a batch size set to 1, which serves as a measure of computational efficiency, indicating the number of floating-point computations required for a single forward (fwd) and backward (bwd) process. We also used trainable parameters (TP) as a metric for comparison.

It can be observed that our method is similar to LoRAMoE in terms of TP and FLOPs. Although TP is slightly higher than that of MoSLoRA, the FLOPs are lower than this method. This demonstrates that the proposed method improves performance without significantly increasing training resources compared to other approaches, and it is considerably more efficient than full fine-tuning.

\section{Related Work}
\subsection{Multimodal Information Extraction}

Multimodal Information Extraction (MIE) encompasses multimodal named entity recognition (MNER), multimodal relation extraction (MRE), and multimodal event extraction (MEE). Previous studies have typically treated these tasks in isolation, overlooking potential correlations between them. 
\newline
\textbf{MNER} addresses the limitations of textual information by incorporating visual data. Early studies enhanced textual representations with visual features \cite{Zhang2018,zhang2021multi,zhang2025mars}. To mitigate the issue of inaccurate image recognition, \cite{yu2020} proposed a unified multimodal transformer that dynamically merges visual and textual representations through a gating mechanism. 
Similarly, \cite{sun2021rpbert} introduced an auxiliary binary classification task to filter irrelevant visual content. Recent approaches, such as ITA \shortcite{wang-etal-2022-ita} and MNER-QG \shortcite{jia2023mner}, enhance unified representations and text-focused attention mechanisms by incorporating image information (e.g., regional object tags) into inputs.
\newline
\textbf{MRE} focuses on extracting relationships between entities and is a newer field compared to MNER \cite{Zheng2021,zheng2021mnre,chen2022hybrid,yuan2023joint,yuan2024fine}. Early methods employed multimodal fusion layers to align textual and visual features, thereby enhancing text representation \cite{chen2022good,zhao2023tsvfn}. However, they often lacked sufficient background information. Recent approaches address this limitation by utilizing retrieval-augmented generation to incorporate relevant images or knowledge from sources such as the Web or Wikipedia, thereby improving relationship identification \cite{yue2023automatic,yuan2024few}
\newline    
\textbf{MEE}
aims to extract events and event parameters from multiple modalities. Unlike MNER and MRE, which solely utilize visual information to enhance text extraction, MEE also extracts visual objects as parameters within events \cite{tong2020image,li2020cross}. 
To better bridge the gap between modalities, \cite{10.1145} proposed a cross-modal contrastive learning framework to improve the similarity between their representations.
\newline 
\textbf{UMIE} 
emerged more recently than unified information extraction (UIE) in text modality. \cite{Sun2024} was the first to propose a unified multimodal information extractor (UMIER), which utilizes the T5 model with a visual adapter to extract both textual and visual information effectively. However, this method utilizes full-parameter fine-tuning combined with instruction tuning, necessitating significant training time and hardware resources. This requirement poses challenges for LLMs and environments with limited hardware resources.

\subsection{Parameter-Efficient Fine-Tuning}

With the rise of LLMs and LVLMs \cite{li2023blip,liu2023visual,kb-vqg,mllm-vcr}, traditional fine-tuning becomes impractical due to the high computational demands of models with billions of parameters \cite{houlsby2019parameter}. Parameter-efficient fine-tuning (PEFT) methods, such as LoRA \shortcite{hu2021lora}, offer a solution by enabling task-specific customization with lower computational cost. LoRA, which applies low-rank updates, maintains the model’s core capabilities, making it widely used for fine-tuning LLMs and LVLMs. Recent studies, such as LoRAMoE \shortcite{dou2024loramoe} and MoSLoRA \shortcite{wu-etal-2024-mixture-subspaces}, have incorporated the Mixture of Experts approach \cite{yuksel2012twenty} into LoRA, leveraging multiple LoRA modules as distinct experts to enhance the model’s ability to utilize world knowledge for solving downstream tasks.

In contrast to vanilla LoRA, LoRAMoE, and MoSLoRA, which adopt single-level approaches, our method addresses gradient conflicts in multi-task instruction fine-tuning through a two-level LoRA framework. This framework consists of a universal LoRA module for shared knowledge and task-specific modules tailored to individual requirements, thereby minimizing conflicts. Additionally, we introduce an achievement-based multi-task loss function that dynamically adjusts based on task performance, helping to mitigate training imbalances across diverse MIE tasks for more balanced and effective training.

\section{Conclusions}
This study proposes a collaborative multi-LoRA experts (C-LoRAE) model that consists of a universal LoRA expert and a set of task-specific LoRA experts to model various MIE tasks jointly. The universal LoRA expert effectively shares knowledge among different instruction tasks, while the task-specific LoRA experts mitigate conflicts arising from diverse instruction data types. Additionally, we designed an achievement-based multi-task loss to address the imbalance caused by the varying numbers of training samples in MIE tasks. This loss function balances the training progress of our proposed C-LoRAE model based on current performance across different MIE instruction tasks. Experiments on seven benchmark datasets across three tasks demonstrate that C-LoRAE not only outperforms traditional full-parameter fine-tuning models and vanilla LoRA fine-tuning methods, achieving state-of-the-art results in four datasets with fewer trainable parameters, highlighting the efficiency and scalability of our method. Future work will explore further optimization of the collaboration among LoRA experts and investigate the application of our approach to other multimodal learning tasks.

\appendix
\section*{Acknowledgments}
This research is supported by the National Natural Science Foundation of China (62476097, 62276072, 62402184), the Fundamental Research Funds for the Central Universities, South China University of Technology (x2rjD2240100),  Guangdong Provincial Fund for Basic and Applied Basic Research—Regional Joint Fund Project (Key Project) (2023B1515120078), Guangdong Provincial Natural Science Foundation for Outstanding Youth Team Project (2024B1515040010), the Hong Kong Polytechnic University under the Postdoc Matching Fund Scheme (Project No. P0049003). Support from the Guangxi Natural Science Foundation Key Project (No. 2025GXNSFDA069017) is also gratefully acknowledged. We further acknowledge the support from the Postdoc Matching Fund Scheme of The Hong Kong Polytechnic University (Project No. P0049003). TW received funding from the NIHR Maudsley Biomedical Research Centre, Maudsley Charity, King’s Together, and MHaPS Early Career Researcher Awards.

\section{More Experiments}
\subsection{The Sensitivity of Hyperparameters}

\begin{figure*}    
  \centering
    \includegraphics[scale=.39]{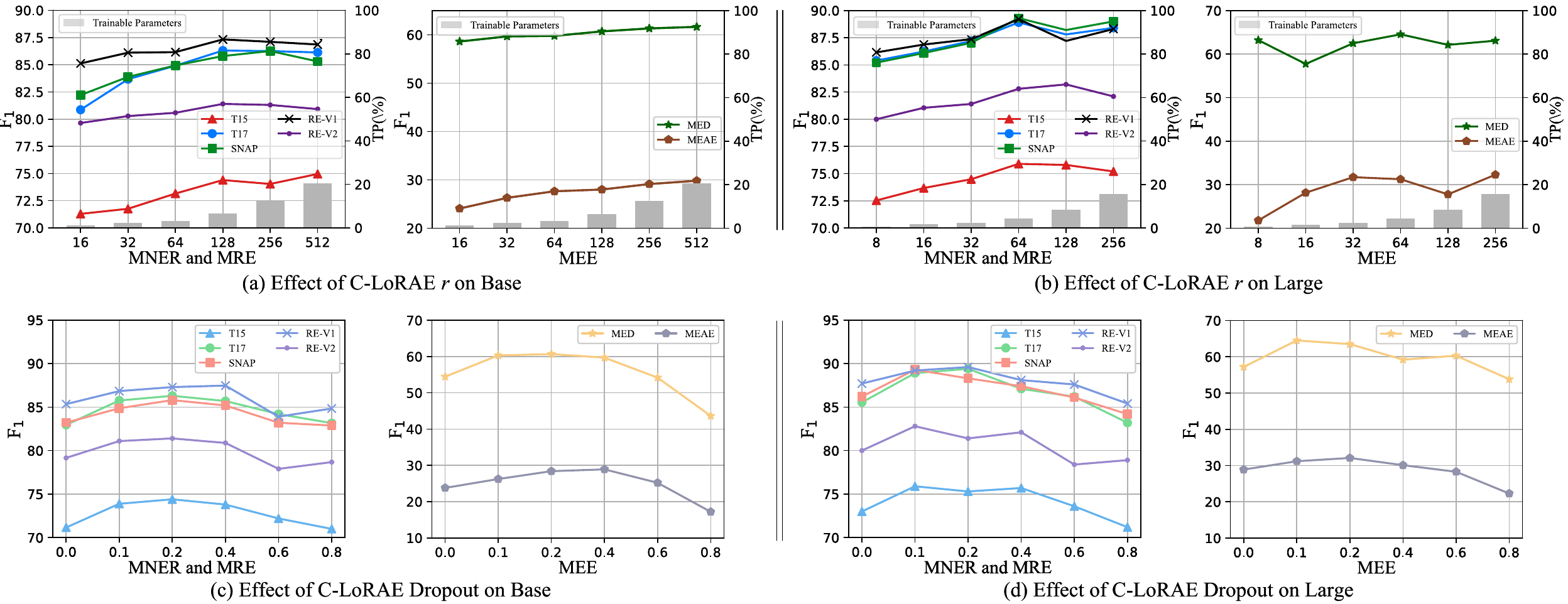}
  \caption{Experimental results of our model with different C-LoRAE parameters. The datasets denoted by T15, T17, RE-V1, and RE-V2 represent Twitter-15, Twitter-17, MRE-V1, and MRE-V2, respectively.}
  \label{FIG:4}
\end{figure*}

To further investigate the impact of each hyperparameter on the final performance of C-LoRAE, we conducted a series of sensitivity experiments. 
In Figures \ref{FIG:4}(a) and (b), for both the \emph{Base} and \emph{Large} models, lower-rank dimensionalities of the applied C-LoRAE fail to support optimal performance in most subtasks. Conversely, larger C-LoRAE configurations may saturate model performance and require exponentially more trainable parameters. Therefore, to balance performance and computational budget, the final dimensions of C-LoRAE are set to 128 and 64 for the \emph{Base} and \emph{Large} models, respectively.

Additionally, we investigated the impact of dropout on C-LoRAE, as shown in Figures \ref{FIG:4}(c) and (d). We found that both the \emph{Base} and \emph{Large} models perform the worst with a dropout rate of 0, indicating that the absence of a dropout reduces the robustness of the model. In contrast, a dropout rate of 0.8 disrupts normal training, leading to poor performance. Optimal performance is achieved when the dropout rate is between 0.1 and 0.4. For the \emph{XLarge} version, we use the same configuration as the \emph{Large} model without further adjusting the relevant parameters.

\subsection{Routing Analysis}
\begin{figure*}    
  \centering
    \includegraphics[scale=.40]{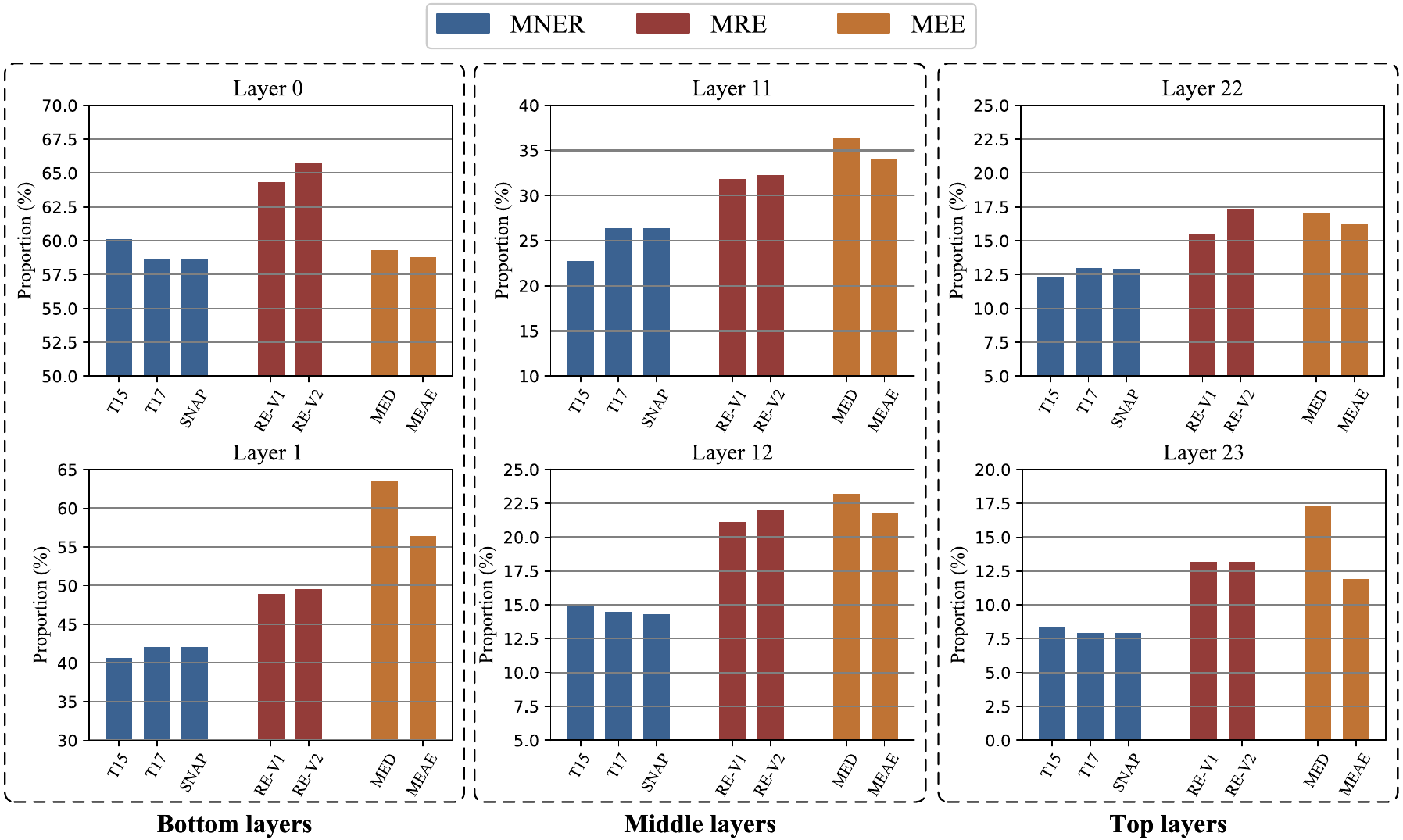}
  \caption{Average proportion of tokens assigned to task-specific experts on different datasets for LLM layers bottom layers(0, 1), middle layers (11, 12), and top layers (23, 24). }
  \label{FIG:layer}
\end{figure*}

To further analyze the role of the expert-motivated gate router between task-specific and universal experts, we visualized the distribution of token proportions allocated to each task-specific LoRA expert across different layers of C-LoRAE$_{Large}$: bottom layers (0, 1), middle layers (11, 12), and top layers (22, 23). The results are shown in Figure \ref{FIG:layer}. Importantly, the sum of proportions for tokens allocated to universal LoRA and task-specific LoRA experts is equal 1.

We observe that, compared to the MNER task, the MRE and MEE tasks tend to rely more on their task-specific LoRA experts, both in the middle and upper layers. These tasks benefit more from independent LoRA experts, which reduces conflicts between instructions. In contrast, the MNER task appears to tend to learn multimodal features through shared knowledge between tasks. Moreover, the routing mechanism of lower layers (0, 1) tends to use task-specific LoRA experts, while the upper layers tend to use universal LoRA experts in both MNER, MRE, and MEE tasks. For some neighboring layers, such as layers 11 and 12, the expert choice patterns are similar across different types of data. In contrast, in certain neighboring layers, such as layers 0 and 1, each type of data exhibits distinct patterns of expert choice.



\bibliographystyle{named}
\bibliography{ijcai25}

\end{document}